\documentclass[sigconf]{acmart}
\usepackage[flushleft]{threeparttable}
\usepackage{xcolor}
\usepackage{amsmath}
\usepackage{multirow}
\usepackage{enumitem}
\usepackage{makecell}
\usepackage{hyperref}

\copyrightyear{2020}
\acmYear{2020}
\setcopyright{acmcopyright} % acmcopyright %rightsretained %acmlicensed
\acmConference[KDD '20] {26th ACM SIGKDD Conference on Knowledge Discovery and Data Mining}{August 23--27, 2020}{Virtual Event, USA}
\acmBooktitle{26th ACM SIGKDD Conference on Knowledge Discovery and Data Mining (KDD '20), August 23--27, 2020, Virtual Event, USA}
\acmPrice{15.00}
\acmDOI{10.1145/3394486.3403314}
\acmISBN{978-1-4503-7998-4/20/08}

\title{AutoFIS: Automatic  Feature Interaction Selection in Factorization  Models for  Click-Through Rate Prediction}
\titlenote{Co-first authors with equal contributions. Bin Liu is now affiliated with ByteDance. Weinan Zhang and Ruiming Tang are the corresponding authors. This work is sponsored by Huawei Innovation Research Program and NSFC (61702327,61772333).}

\author{Bin Liu$^{\dag*}$, Chenxu Zhu$^{\ddag*}$, Guilin Li$^{\dag*}$, Weinan Zhang$^{\ddag}$\\
	\vspace{-13pt} Jincai Lai$^{\dag}$, Ruiming Tang$^{\dag}$, Xiuqiang He$^{\dag}$, Zhenguo Li$^{\dag}$, Yong Yu$^{\ddag}$}%
\affiliation{$^{\dag}$Huawei Noah's Ark Lab~~~$^{\ddag}$Shanghai Jiao Tong University\\
liubinbh@126.com; ~~\{zhuchenxv,wnzhang\}@sjtu.edu.cn;  \{liguilin2,tangruiming\}@huawei.com}

\begin{document}
\fancyhead{}

\begin{abstract}
Learning feature interactions is crucial for click-through rate (CTR) prediction in recommender systems. In most existing deep learning models, feature interactions are either manually designed or simply enumerated.  However, enumerating all feature interactions  brings large memory and computation cost. Even worse, useless interactions may introduce noise and complicate the training process. In this work, we propose a two-stage algorithm called Automatic Feature Interaction Selection (AutoFIS).  AutoFIS can automatically identify important feature interactions for  factorization models with computational cost just equivalent to  training the target model to convergence. In the \emph{search stage}, instead of searching over a discrete set of candidate feature interactions, we relax the choices to be continuous by introducing the architecture parameters. By implementing  a regularized optimizer over the architecture parameters,  the model can automatically identify and remove the redundant feature interactions during the training process of the model.  In the \emph{re-train stage}, we keep the architecture parameters serving as an attention unit to further boost the performance.    Offline experiments on three large-scale datasets (two public benchmarks, one private) demonstrate that AutoFIS can significantly improve   various FM based models. AutoFIS has been deployed onto the training platform of Huawei App Store recommendation service, where a 10-day online A/B test demonstrated that AutoFIS improved the DeepFM model by 20.3\% and 20.1\% in terms of CTR and CVR respectively.

\end{abstract}

\renewcommand{\shortauthors}{}
\renewcommand{\shorttitle}{AutoFIS: Automatic  Feature Interaction Selection}




\maketitle

%\vspace{-0.5em}
{\fontsize{8pt}{8pt} \selectfont \textbf{ACM Reference Format:} \\
Biu Liu, Chenxu Zhu, Guilin Li, Weinan Zhang, Jincai Lai, Ruiming Tang, Xiuqiang He, Zhenguo Li, Yong Yu. 2020. AutoFIS: Automatic  Feature Interaction Selection in Factorization  Models for  Click-Through Rate Prediction. In \textit{26th ACM SIGKDD Conference on Knowledge Discovery and Data Mining (KDD'20), August 23--27, 2020, Virtual Event, USA. } ACM, New York, NY, USA, 10 pages. https://doi.org/10.1145/3394486.3403314}

% \footnote{$*$These authors contributed equally to this research.}

\section{Introduction}
\label{sec:intro}
Click-through rate (CTR) prediction is crucial in recommender systems, where the task is to predict the probability of the user clicking on the recommended items (e.g., movie, advertisement)~\cite{widedeep,din}. Many recommendation decisions can then be made based on the predicted CTR. 
The core of these recommender systems is to extract significant low-order and high-order feature interactions. 

Explicit feature interactions can significantly improve the performance of CTR models~\cite{deepfm, ftrl, gbdtlr, latent-cross}. Early collaborative filtering recommendation algorithms, such as matrix factorization (MF)~\cite{mf} and factorization machine (FM)~\cite{fm}, extract second-order information with a bi-linear learning model.

However, not all interactions are conducive to performance. Some tree-based methods have been proposed to find useful intersections automatically. Gradient boosting decision tree (GBDT)~\cite{gbdt, gbdtlr} tries to find the interactions with higher gradients of the loss function. AutoCross~\cite{autocross} searches effective interactions in a tree-structured space. But tree models can only explore a small fraction of all possible feature interactions in recommender systems with multi-field categorical data~\cite{pin}, so their exploration ability is restricted.

In the meantime, Deep Neural Network (DNN) models~\cite{dnnyoutube,zhang2016deep} are proposed. Their representational ability is stronger and they could explore most of the feature interactions according to the universal approximation property~\cite{DNN_universial}. However, there is no guarantee that a DNN naturally converges to any expected functions using gradient-based optimization.
A recent work proves the insensitive gradient issue of DNN when the target is a large collection of uncorrelated functions~\cite{fail_dnn,pin}. Simple DNN models may not find the proper feature interactions.
Therefore, various complicated architectures have been proposed, such as Deep Interest Network (DIN)~\cite{din}, Deep Factorization Machine (DeepFM)~\cite{deepfm}, Product-based Neural Network (PNN)~\cite{Qu2016Product,pin}, and Wide \& Deep~\cite{widedeep}. \textbf{Factorization Models} (specified in Definition 1), such as FM, DeepFM, PNN, Attention Factorization Machine (AFM) ~\cite{afm}, Neural Factorization Machine (NFM)~\cite{nfm}, have been proposed to adopt a feature extractor to explore explicit feature interactions.

% However, not all interactions are conducive to performance. Some tree-based methods have been proposed to find useful intersections automatically. Gradient boosting decision tree (GBDT)~\cite{gbdt, gbdtlr} transform input features and extract important interactions. AutoCross~\cite{autocross} search efficient interactions in a tree-structured space.

% Due to substantial computational costs, all these models select interactions only by decision tree models or deep neural network (DNN) models. Because of the sparsity issue of trees and insensitive gradient issue of DNN, training feature interactions is difficult and the interactions are hard to extract all useful information~\cite{pin}.  
% Otherwise, the information extracted by these interactions is more suitable for simple models, such as logistic regression (LR)~\cite{lr, Ren2016User}, wide \& deep model~\cite{widedeep}, field-aware factorization machine (FFM)~\cite{ffm} and not suitable to transfer to more complicated deep models.

% In the meantime, various complicated deep methods~\cite{din, deepfm, pin} have been proposed, such as DeepFM and PNN. DeepFM, and PNN adopt a feature extractor to explore field-aware feature interactions. In this way, these models could extract information not found in DNN models. 

However, all these models are simply either enumerating all feature interactions or requiring human efforts to identify important feature interactions. The former always brings large memory and computation cost to the model and is difficult to be extended into high-order interactions. Besides, useless interactions may bring unnecessary noise and complicate the training process~\cite{fail_dnn}. The latter, such as identifying important interactions manually in Wide \& Deep~\cite{widedeep}, is of high labor cost and risks missing some counter-intuitive (but important) interactions.

If useful feature interactions can be identified beforehand in these factorization models, the models can focus on learning over them without having to deal with useless feature interactions. Through removing the useless or even harmful interactions, we would expect the model to perform better with reduced computation cost.

To automatically learn which feature interactions are essential,  we introduce a \textit{gate} (in open or closed status) for each feature interaction to control whether its output should be passed to the next layer. In previous works, the status of the gates is either specified beforehand by expert knowledge ~\cite{widedeep} or set as all open~\cite{deepfm,xdeepfm}. From a data-driven point of view,  whether open or closed a gate should depend on the contribution of each feature interaction to the final prediction. Apparently, those contributing little should be closed to prevent introducing extra noise to model learning. However, it is an NP-Hard problem to find the optimal set of open gates for model performance, as we face an incredibly huge ($2^{\mathcal{C}_{m}^{2}}$, with $m$ the number of feature fields, if we only consider $2^{nd}$-order feature interactions) space to search. 

%Some researchers propose a method called AutoCross~\cite{luo2019autocross} to automatically select feature interactions in a interaction tree. Although it reduces complexity to some extent, it costs so much that it is unacceptable in most applications. 

Inspired by the recent work DARTS~\cite{liu2018darts,MiLeNAS,li2019stacnas} for neural architecture search, we propose a two-stage method AutoFIS for automatic selecting low-order and high-order feature interactions in \emph{factorization models}.
%(defined in Definition 1).
% , such as FM~\cite{fm}, DeepFM~\cite{deepfm}, IPNN~\cite{pin}, AFM~\cite{afm}, NFM~\cite{nfm}.
In the \emph{search stage}, instead of searching over a discrete set of candidate feature interactions, we relax the choices to be continuous by introducing a set of architecture parameters (one for each feature interaction) so that the relative importance of each feature interaction can be learned by gradient descent. 
The architecture parameters are jointly optimized with neural network weights by GRDA optimizer~\cite{chao2019generalization} (an optimizer which is easy to produce a sparse solution) so that the training process can automatically abandon unimportant feature interactions (with zero values as the architecture parameters) and keep those important ones.
After that, in the \emph{re-train stage}, we select the feature interactions with non-zero values of the architecture parameters and re-train the model with the selected interactions while keeping the architecture parameters as attention units instead of indicators of interaction importance.
% The architecture parameters are jointly optimized with neural network weights so that the training process can automatically assign higher values of the architecture parameter to those more important feature interactions. 
% After that, in the \emph{re-train stage}, we select a subset of feature interactions with higher values of the architecture parameter and re-trains the model with the selected interactions while keeping the architecture parameters as attention units instead of indicators of interaction importance. 
%To obtain the final architecture, we keep the architecture parameters together with the selected subset of feature interactions, where the set of architecture parameters now serve as attention units, instead of indicators for selection decisions. 

Extensive experiments are conducted on three large-scale datasets (two are public benchmarks, and the other is private). Experimental results demonstrate that AutoFIS can significantly improve the CTR prediction performance of factorization models on all datasets. As AutoFIS can remove about 50\%-80\%  $2^{nd}$-order feature interactions, original models can always achieve improvement on efficiency. We also apply AutoFIS for $3^{rd}$-order interaction selection by learning the importance of each $3^{rd}$-order feature interaction. Experimental results show that with about 1\%--10\% of the $3^{rd}$-order interactions selected, the AUC of factorization models can be improved by 0.1\%--0.2\% without introducing much computation cost. The results show a promising direction of using AutoFIS for automatic high-order feature interaction selection.  Experiments also demonstrate that important $2^{nd}$- and $3^{rd}$-order feature interactions, identified by AutoFIS in factorization machine, can also greatly boost the performance of current state-of-the-art models, which means we can use a simple model for interaction selection and apply the selection results to other models. 
Besides, we analyze the effectiveness of feature interactions selected by our model on real data and synthetic data.
Furthermore, a ten-day online A/B test is performed in a Huawei App Store recommendation service, where AutoFIS yielding recommendation model achieves improvement of CTR by $20.3\%$ and CVR by $20.1\%$ over DeepFM, which contributes a significant business revenue growth.

To summarize, the main contributions of this paper can be highlighted as follows:
\begin{enumerate}[leftmargin = 10 pt]
\item We empirically verify that removing the redundant feature interactions is beneficial when training factorization models.
% \item We are the first to empirically prove that removing the redundant feature interactions is beneficial when training factorization models.  
\item We propose a two-stage algorithm \textbf{AutoFIS} to automatically select important low-order and high-order feature interactions in factorization models. In the \emph{search stage}, AutoFIS can learn the relative importance of each feature interaction via architecture parameters within one full training process.
%we formulate the relative importance of different feature interactions as a set of architecture parameters, which can be jointly trained with the neural network weights. This allows us to score every interaction with one full training of the architecture.
In the \emph{re-train stage}, with the unimportant interactions removed, we re-train the resulting neural network while keeping architecture parameters as attention units to help the learning of the model. 

\item Offline experiments on three large-scale datasets demonstrate the superior performance of AutoFIS in factorization models.  Moreover, AutoFIS can also find a set of important high-order feature interactions to boost the performance of existing models without much computation cost introduced. A ten-day online A/B test shows that AutoFIS improves DeepFM model by approximately 20\% on average in terms of CTR and CVR.

% \item By using AutoFIS, we are able to include the $3^{rd}$-order feature interactions in the model with the same inference cost as the original FM model, which further improves the model accuracy. Taking Avazu dataset as an example, when using AutoFIS to select both second and third order feature interactions，the relative improvement over FM and DeepFM is  0.8\% and 0.4\% respectively, while the inference cost is almost the same.   
% %\item Offline evaluation on three large-scale datasets (two are public benchmark, and the other is private) demonstrate AutoFIS can significantly improve the accuracy of FM based models. Online experiments are conducted to show that AutoFIS improves DeepFM model by 20\% in average in terms of CTR and CVR in a ten-day AB test.

\end{enumerate}

\begin{figure*}
    \centering
    \vspace{-1em}
    \includegraphics[width=0.9 \textwidth]{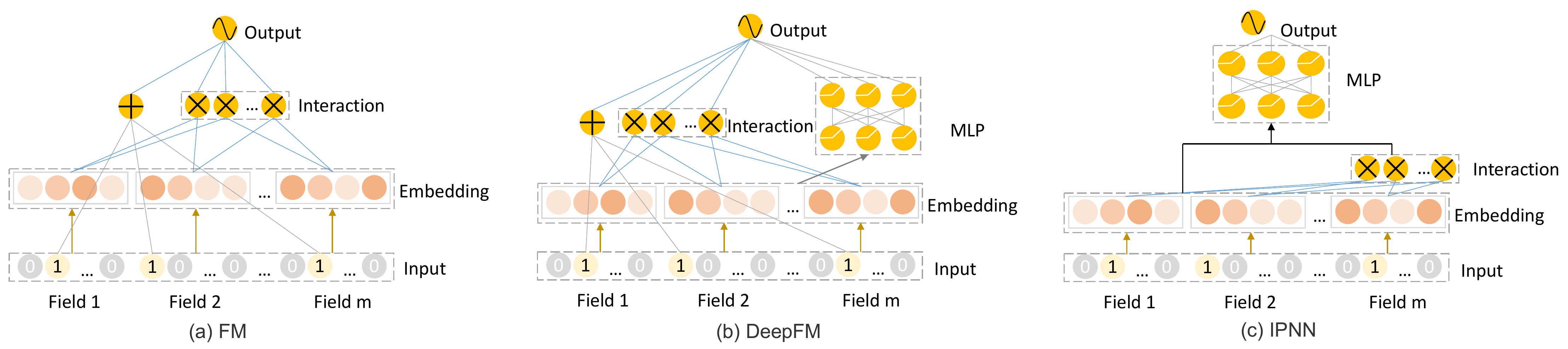}
    \vspace{-1.5em}
    \caption{\small Architectures of FM, DeepFM and IPNN}
    \vspace{-1.4em}
    \label{fig:fm_deepfm_ipnn}
\end{figure*}

\section{Related Work}
CTR prediction is generally formulated as a binary classification problem~\cite{fgcnn}. In this section we briefly review factorization models for CTR prediction and AutoML models for recommender systems.

Factorization machine (FM)~\cite{fm} projects each feature into a low-dimensional vector and models feature interactions by inner product, which works well for sparse data. Field-aware factorization machine (FFM) \cite{ffm} further enables each feature to have multiple vector representations to interact with features from other fields. 

Recently, deep learning models have achieved state-of-the-art performance on some public benchmarks~\cite{fgcnn,crossnet}. Several models use MLP to improve FM, such as Attention FM~\cite{afm}, Neural FM~\cite{nfm}. Wide \& Deep ~\cite{widedeep} jointly trains a wide model for artificial features and a deep model for raw features.  DeepFM~\cite{deepfm} uses an FM layer to replace the wide component in Wide \& Deep. PNN~\cite{Qu2016Product} uses MLP to model  the interaction  of FM layer and feature embeddings while 
PIN~\cite{pin} introduces a network-in-network architecture to model pairwise feature interactions with sub-networks rather than simple inner product operations in PNN and DeepFM. Note that all existing factorization models simply enumerate all $2^{nd}$-order feature interactions which contain many useless and noisy interactions. 

Gradient boosting decision tree (GBDT)~\cite{gbdt} is a method to do feature engineering and search interactions by decision tree algorithm.
Then the transformed feature interactions can be fed into to logistic regression \cite{gbdtlr} or FFM \cite{juan2014idiots}. In practice, tree models are more suitable for continuous data but not for high-dimensional categorical data in recommender system because of the low usage rate of categorical features \cite{pin}.

In the meantime, there exist some works using AutoML techniques to deal with the problems in recommender system. 
AutoCross~\cite{autocross} is proposed to search over many subsets of candidate features to identify effective interactions. This requires training the whole model to evaluate the selected feature interactions, but the candidate sets are incredibly many: i.e.,  there are $2^{\mathcal{C}_{m}^{2}}$ candidate sets for a dataset with $m$ fields for just $2^{nd}$-order feature interactions. Thus AutoCross accelerates by two aspects of approximation: (i) it greedily constructs local-optimal feature sets via beam search in a tree structure, and (ii) it evaluates the newly generated feature sets via field-aware logistic regression. Due to such two approximations, the high-order feature interactions extracted from AutoCross may not be useful for deep models. Compared with AutoCross, our proposed AutoFIS only needs to perform the search stage once to evaluate the importance of all feature interactions, which is much more efficient. Moreover, the learned useful interactions will improve the deep model as they are learned and evaluated in this deep model directly.

Recently, one-shot architecture search methods, such as DARTS \cite{liu2018darts}, have become the most popular neural architecture search (NAS) algorithms to efficiently search network architectures \cite{bender2019understanding}.  
In recommender systems, such methods are utilized to search proper interaction functions for collaborative filtering models~\cite{autoML_for_CF}. 
The model in~\cite{autoML_for_CF} focuses on identifying proper interaction functions for feature interactions while our model focuses on searching and keeping important feature interactions.
Inspired by the recent work DARTS for neural architecture search, we formulate the problem of searching the effective feature interactions as a continuous searching problem by incorporating architecture parameters. Different from DARTS using two-level optimization to optimize the architecture parameters and the model weights alternatively and iteratively with the training set and the validation set,  we use one-level optimization to train these two types of parameters jointly with all data as the training set. We analyze their difference theoretically in Section \ref{sec:autofis}, and compare their performance in the Experiments Section. 
% note:  someone may confuse that we use two level optimization because alpha and w we use different optimizer. But the difference between two-level and one-level should be whether a validation set is used to update alpha. 

\section{METHODOLOGY}
In this section, we describe the proposed AutoFIS, an algorithm to select important feature interactions in factorization models automatically.

\subsection{Factorization Model (Base Model)}
First, we define factorization models:
\begin{definition}
\textbf{Factorization models} are the models where the interaction of several embeddings from different features is modeled into a real number by some operation such as inner product or neural network.
\end{definition}

We take FM, DeepFM, and IPNN as instances to formulate our algorithm and explore the performance on various datasets. 
Figure~\ref{fig:fm_deepfm_ipnn} presents the architectures of FM, DeepFM and IPNN models. FM consists of a \emph{feature embedding layer} and a \emph{feature interaction layer}. Besides these two layers, DeepFM and IPNN model include an extra layer: \emph{MLP layer}. The difference between DeepFM and IPNN is that feature interaction layer and MLP layer work in parallel in DeepFM, while ordered in sequence in IPNN.

In the subsequent subsections, we will brief the feature embedding layer and feature interaction layer in FM. To work with DeepFM and IPNN model, the MLP layer and output layer are also formalized. Then the detail of how our proposed AutoFIS working on the feature interaction layers is elaborated, i.e., selecting important feature interactions based on architecture parameters. 

\textbf{Feature Embedding Layer.}
 In most CTR prediction tasks, data is collected in a multi-field categorical form\footnote{Features in numerical form are usually transformed into categorical form by bucketing.}. A typical data pre-process is to transform each data instance into a high-dimensional sparse vector via one-hot or multi-hot encoding. 
A field is represented as a multi-hot encoding vector only when it is multivariate.
A data instance can be represented as \[\textbf{\textit{x}} = [\textbf{\textit{x}}_1, \textbf{\textit{x}}_2, \cdots, \textbf{\textit{x}}_m],\]
% \[\textit{x} = [ \textit{x}_{1}, \textit{x}_{2}, ..., \textit{x}_{m}],\]
where $m$ is the number of fields and $\boldsymbol{x}_i$ is the one-hot or multi-hot encoding vector of the $i$-th field. A feature embedding layer is used to transform an encoding vector into a low-dimensional vector as  \begin{equation}
    \textbf{\textit{e}}_i = V_i \textbf{\textit{x}}_i.
\end{equation}
where $V_i\in R^{d\times n_i}$ is the a matrix, $n_i$ is the number of feature values in the $i$-th field and $d$ is the dimension of low-dimensional vectors.
\begin{itemize}
    \item If $\textbf{\textit{x}}_i$ is a one-hot vector with $j$-th element $\textbf{\textit{x}}_i[j]=1$, then the representation of $\textbf{\textit{x}}_i$ is $V_i^j$.
    \item If $\textbf{\textit{x}}_i$ is a multi-hot vector with $\textbf{\textit{x}}_i[j]=1$ for $j=i_1,i_2,\cdots,i_k$ and the embeddings of these elements are $\{V_i^{i1},V_i^{i2},\cdots,V_i^{ik}\}$, then the representation of $\textbf{\textit{x}}_i$ is the sum or average of these embeddings~\cite{dnnyoutube}.
\end{itemize}

% % A feature embedding layer is used to transform each one-hot encoding vector into a low-dimensional vector as
% \[\textit{e}_{i}=\textit{V}_{i}\textit{x}_{i},\] where $\textit{V}_{i}$ is the embedding matrix for all the features in field $i$ and $\textit{e}_{i}$ is the embedding vector of the feature in the $i$-th field of \textit{x}, retrieving from $\textit{V}_{i}$ with $\textit{x}_{i}$ as the index. If a field is multivariate, it is represented as multi-hot encoding vectors and its embedding takes the sum or average of their embeddings~\cite{youtube-dnn}.

The output of the  feature embedding layer is then the concatenation of multiple embedding vectors as
\[
\textbf{\textit{E}} = [\textbf{\textit{e}}_1, \textbf{\textit{e}}_2, ..., \textbf{\textit{e}}_m].
\]

\textbf{Feature Interaction Layer.}
 After transforming the features to low-dimensional space,  the feature interactions can be modeled in such a space with the feature interaction layer. First, the inner product of the pairwise feature interactions is calculated:
\begin{equation}
[\langle\textbf{\textit{e}}_{1}, \textbf{\textit{e}}_{2}\rangle, \langle\textbf{\textit{e}}_{1}, \textbf{\textit{e}}_{3}\rangle, ..., \langle\textbf{\textit{e}}_{m-1}, \textbf{\textit{e}}_{m}\rangle],
\end{equation}
where $\textbf{\textit{e}}_{i}$  is the feature embedding of $i$-th field, $\langle\cdot , \cdot\rangle$ is the inner product of two vectors. The number of pair-wise feature interactions in this layer is $\mathcal{C}_{m}^{2}$.

In FM and DeepFM models, the output of the feature interaction layer is: 
%the summation of an Addition unit and the feature interaction units:
\begin{equation}
    l_{fm}=\langle \boldsymbol{w},\boldsymbol{x}\rangle +\sum_{i=1}^m\sum_{j>i}^m \langle \textbf{\textit{e}}_{i}, \textbf{\textit{e}}_{j}\rangle.
    \label{eqn:FM_inter1}
\end{equation}

Here, all the feature interactions are passed to the next layer with equal contribution. As pointed in Section~\ref{sec:intro} and will be verified in Section~\ref{sec:exp}, not all the feature interactions are equally predictive and useless interactions may even degrade the performance. Therefore, we propose the AutoFIS algorithm to select important feature interactions efficiently.

To study whether our methods can be used to identify important high-order interactions, we define the feature interaction layer with  $3^{rd}$-order interactions (i.e., combination of three fields) as:
\begin{equation}
    l_{fm}^{3rd}=\langle \boldsymbol{w},\boldsymbol{x}\rangle +\sum_{i=1}^m\sum_{j>i}^m \langle\textbf{\textit{e}}_{i}, \textbf{\textit{e}}_{j}\rangle + \sum_{i=1}^m\sum_{j>i}^m \sum_{t>j}^m\langle\textbf{\textit{e}}_{i}, \textbf{\textit{e}}_{j}, \textbf{\textit{e}}_{t}\rangle.
    \label{eqn:FM_inter}
\end{equation}

\textbf{MLP Layer.} 
 MLP Layer consists of several fully connected layers with activation functions, which learns the relationship and combination of features. The output of one such layer is
    \begin{equation}\label{eqn:MLP}
        \boldsymbol{a}^{(l+1)}=\mbox{relu}(W^{(l)}\boldsymbol{a}^{(l)}+\boldsymbol{b}^{(l)}),
    \end{equation}
where $\boldsymbol{a}^{(l)}, W^{(l)}, \boldsymbol{b}^{(l)}$ are the input, model weight, and bias of the $l$-th layer. Activation $\mbox{relu}(z)=\max(0,z)$. $\boldsymbol{a}^{(0)}$ is the input layer and $\texttt{MLP}(\boldsymbol{a}^{(0)}) = \boldsymbol{a}^{(h)}$, where $h$ is the depth of MLP layer \texttt{MLP}.
    
\textbf{Output Layer.}
 FM model has no MLP layer and connects the feature interaction layer with prediction layer directly:
\begin{equation}
    \hat{y}_{\texttt{FM}} = \mbox{sigmoid}( l_{fm} ) = \frac{1}{1 + \exp (-l_{fm})},
\end{equation}
where $\hat{y}_{\texttt{FM}} $ is the predicted CTR. 

DeepFM combines feature interaction layer and MLP layers in parallel as 
\begin{equation}
    \hat{y}_{\texttt{DeepFM}} = \mbox{sigmoid}(l_{fm}+ \texttt{MLP}(\boldsymbol{E})).
    \label{eqn:deepfm}
\end{equation}
While in IPNN, MLP layer is sequential to feature interaction layer as 
\begin{equation}
 \hat{y}_{\texttt{IPNN}} = \mbox{sigmoid}(\texttt{MLP}([\boldsymbol{E},l_{fm}])).   
\end{equation}

Note that the MLP layer of IPNN can also serve as a re-weighting of the different feature interactions, to capture their relative importance. This is also the reason that IPNN has a higher capacity than FM and DeepFM. However, with the IPNN formulation, one cannot retrieve the exact value corresponding to the relative contribution of each feature interaction. Therefore, the useless feature interactions in IPNN can be neither identified nor dropped, which brings extra noise and computation cost to the model. We would show in the following subsections and Section~\ref{sec:exp} that how the proposed method AutoFIS could improve IPNN. 

\textbf{Objective Function.} 
 FM, DeepFM, and IPNN share the same objective function,  i.e., to minimize the cross-entropy of predicted values and the labels as 
\begin{equation}
    \mathcal{L}(y,\hat{y}_{\texttt{M}}) = -y\mbox{log}\hat{y}_{\texttt{M}}- (1-y)\mbox{log}(1-\hat{y}_{\texttt{M}}),
    \label{eqn:loss}
\end{equation}
where $y\in\{0,1\}$ is the label and $\hat{y}_{\texttt{M}} \in [0,1]$ is the predicted probability of $y=1$. 

\subsection{AutoFIS}\label{sec:autofis}
\begin{figure}
    \centering
    \vspace{-0.2em}
    \includegraphics[width=0.38\textwidth]{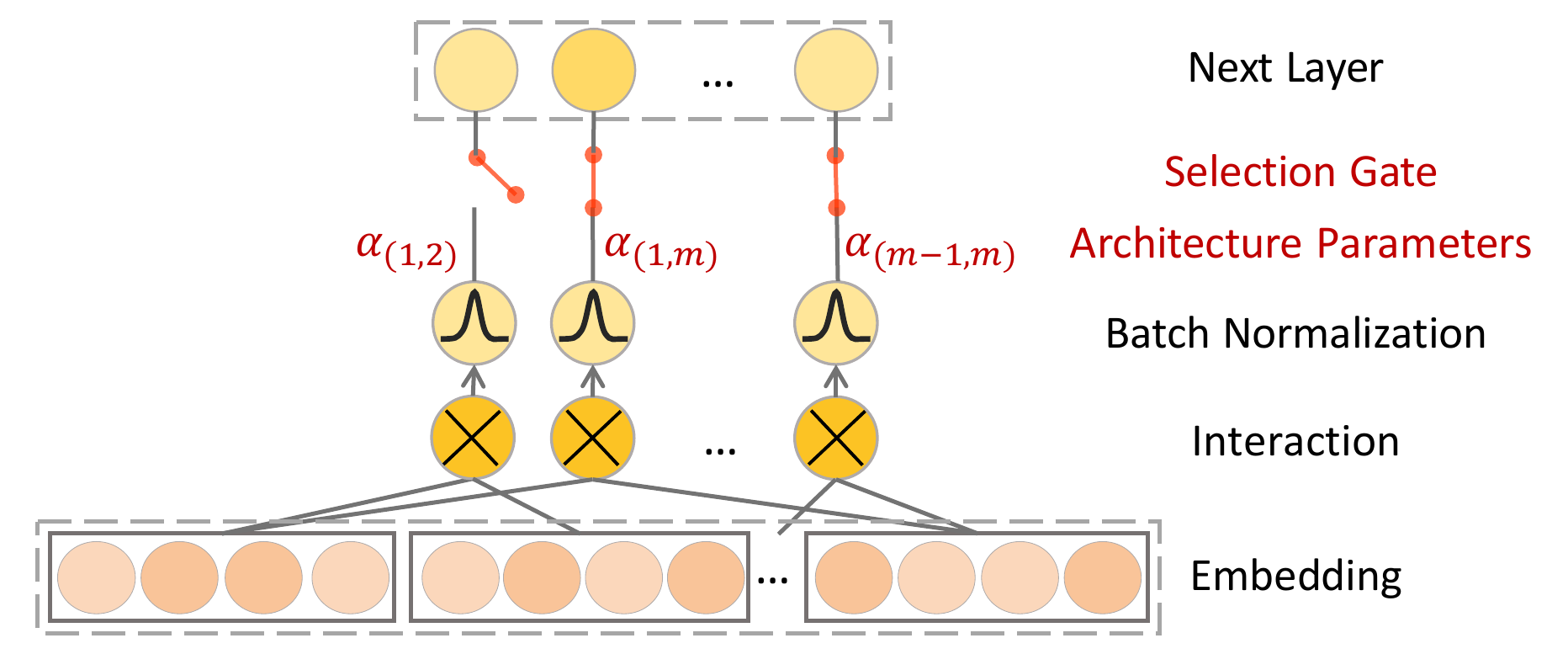}
    \vspace{-1.3em}
    \caption{\small Overview of AutoFIS}
    \vspace{-1.8em}
    \label{fig:autoFIS}
\end{figure}
AutoFIS automatically selects useful feature interactions, which can be applied to the feature interaction layer of any factorization model. In this section, we elaborate on how it works. AutoFIS can be split into two stages: \emph{search stage} and \emph{re-train stage}. In the search stage, AutoFIS detects useful feature interactions; while in the re-train stage, the model with selected feature interactions is re-trained.

\textbf{Search Stage.} 
To facilitate the presentation of the algorithm, we introduce the  \emph{gate} operation to control whether to select a feature interaction: an open gate corresponds to selecting a feature interaction, while a closed gate results in a dropped interaction. The total number of gates corresponding to all the $2^{nd}$-order feature interactions is $\mathcal{C}_{m}^{2}$. It is very challenging to find the optimal set of open gates in a brute-force way, as we face an incredibly huge ($2^{\mathcal{C}_{m}^{2}}$) space to search. In this work, we approach the problem from a different viewpoint: instead of searching over a discrete set of open gates, we relax the choices to be continuous by introducing architecture parameters $\boldsymbol{\alpha}$, so that the relative importance of each feature interaction can be learned by gradient descent. The overview of the proposed AutoFIS is illustrated in Figure~\ref{fig:autoFIS}.

This architecture selection scheme by gradient learning is inspired by DARTS \cite{liu2018darts}, where the objective is to select one operation from a set of candidate operations in convolutional neural network (CNN) architecture. 

To be specific, we reformulate the interaction layer in factorization models (shown in Equation~\ref{eqn:FM_inter1}) as
\begin{equation}
    l_{\texttt{AutoFIS}}  = \langle w,x\rangle +\sum_{i=1}^m\sum_{j>i}^m \alpha_{(i,j)}\langle\textbf{\textit{e}}_{i}, \textbf{\textit{e}}_{j}\rangle,
    \label{eqn:FM_autointer}
\end{equation}
where $\boldsymbol{\alpha}=\{\alpha_{(1,2)}, \cdots, \alpha_{(m-1,m)}\}$ are the architecture parameters.
In the \emph{search stage} of AutoFIS, $\alpha_{(i,j)}$ values are learned in such a way that $\alpha_{(i,j)}$ can represent the relative contribution of each feature interaction to the final prediction. Then, we can decide the gate status of each feature interaction by setting those unimportant ones (i.e., with zero $\alpha_{(i,j)}$ values) closed. 

\textbf{Batch Normalization.}
 From the viewpoint of the overall neural network, the contribution of a feature interaction is measured by $\alpha_{(i,j)}\cdot \langle \textbf{\textit{e}}_i, \textbf{\textit{e}}_j\rangle$ (in Equation~\ref{eqn:FM_autointer}). Exactly the same contribution can be achieved by re-scaling this term as  $(\frac{\alpha_{(i,j)}}{\eta})\cdot (\eta\cdot \langle \textbf{\textit{e}}_i, \textbf{\textit{e}}_j\rangle)$, where $\eta$ is a real number.

Since the value of $\langle \textbf{\textit{e}}_i, \textbf{\textit{e}}_j\rangle$ is jointly learned with $\alpha_{(i,j)}$, the coupling of their scale would lead to unstable estimation of $\alpha_{(i,j)}$, such that $\alpha_{(i,j)}$ can no longer represent the relative importance of $\langle \textbf{\textit{e}}_i, \textbf{\textit{e}}_j\rangle$. 
To solve this problem, we apply Batch Normalization (BN)~\cite{bn} on $\langle \textbf{\textit{e}}_i, \textbf{\textit{e}}_j\rangle$ to eliminate its scale issue. BN has been adopted by training deep neural networks as a standard approach to achieve fast convergence and better performance. The way that BN normalizes values gives an efficient yet effective way to solve the coupling problem of $\alpha_{(i,j)}$ and $\langle \textbf{\textit{e}}_i, \textbf{\textit{e}}_j\rangle$. 

The original BN normalizes the activated output with statistics information of a mini-batch. Specifically, 
\begin{equation}
\hat{\boldsymbol{z}} = \frac{\boldsymbol{z}_{in}-\boldsymbol{\mu}_{\mathcal{B}}}{\sqrt{
\sigma_{\mathcal{B}}^{2}+\epsilon}} \hspace{10pt}\text{and}\hspace{10pt} \boldsymbol{z}_{out} = \theta \cdot \hat{\boldsymbol{z}} + \boldsymbol{\beta},    
\end{equation}
where $\boldsymbol{z}_{in}$, $\hat{\boldsymbol{z}}$ and $\boldsymbol{z}_{out}$ are input, normalized and output values of BN; $\boldsymbol{\mu}_{\mathcal{B}}$ and $\sigma_{\mathcal{B}}$ are the mean and standard deviation values of $\boldsymbol{z}_{in}$ over a mini-batch $\mathcal{B}$; $\theta$ and $\boldsymbol{\beta}$ are trainable \emph{scale} and \emph{shift} parameters of BN; $\epsilon$ is a constant for numerical stability.
% \weinan{explain $\epsilon$.}

To get stable estimation of $\alpha_{(i,j)}$, we set the \emph{scale} and \emph{shift} parameters to be 1 and 0 respectively. The BN operation on each feature interaction $\langle \textbf{\textit{e}}_i,\textbf{\textit{e}}_j\rangle$ is calculated as 
\begin{equation}
\langle \textbf{\textit{e}}_i,\textbf{\textit{e}}_j\rangle_{BN} = \frac{\langle \textbf{\textit{e}}_i,\textbf{\textit{e}}_j\rangle - \mu_{\mathcal{B}}(\langle \textbf{\textit{e}}_i,\textbf{\textit{e}}_j\rangle)}{\sqrt{
\sigma_{\mathcal{B}}^{2}(\langle \textbf{\textit{e}}_i,\textbf{\textit{e}}_j\rangle)+\epsilon}},
\end{equation}
where $\mu_{\mathcal{B}}$ and $\sigma_{\mathcal{B}}$ are the mean and standard deviation of $\langle \textbf{\textit{e}}_i,\textbf{\textit{e}}_j\rangle$ in mini-batch $\mathcal{B}$.

\textbf{GRDA Optimizer}.
 Generalized  regularized dual averaging (GRDA) optimizer~\cite{chao2019generalization,chengguang2} is aimed to get a sparse  deep neural network. 
% without necessarily sacrificing testing accuracy. 
To update $\alpha$ at each gradient step $t$ with data $Z_{t}$ we use the following equation:  
\begin{equation}
\footnotesize
    \alpha_{t+1} = \arg \min \limits_\alpha \{\alpha^T(-\alpha_0+\gamma \sum_{i=0}^t \nabla  L(\alpha_t;Z_{i+1})+g(t,\gamma)\|\alpha\|_1+1/2\|\alpha\|_2^2\}
\end{equation}
where $g(t,\gamma)=c \gamma^{1/2}(t\gamma)^u$, and $\gamma$ is  the learning rate,  $c$ and $\mu$ are adjustable hyper-parameters to trade-off between accuracy and sparsity.

%If one starts from a dense network, GRDA optimizer can produce a sparse solution without sacrificing much accuracy.
%If one starts from a dense network, GRDA optimizer can identify the redundant part and  produce a sparse solution.
% Note: in our case, the acc is improved. 

In the search stage, we use GRDA optimizer to learn the architecture parameters $\boldsymbol{\alpha}$ and get a sparse solution. Those unimportant feature interactions (i.e., with zero $\alpha_{(i,j)}$ values) will be thrown away automatically. Other parameters are learned by Adam optimizer, as normal.

% \weinan{After reading this subsubsection, I still don't know how GRDA works. I suggest to add a formula of it.}

\textbf{One-level Optimization.} 
To learn the architecture parameters $\alpha_{(i,j)}$ in the \emph{search stage} of AutoFIS, we propose to optimize $\boldsymbol{\alpha}$ jointly with all the other network weights $\boldsymbol{v}$ (such as $\boldsymbol{w}$ in Equation~\ref{eqn:FM_inter1} and $W^{(l)}, \boldsymbol{b}^{(l)}$ in Equation \ref{eqn:MLP}). This is different from DARTS. DARTS treats $\boldsymbol{\alpha}$ as higher-level decision variables and the network weights as lower-level variables, then optimizes them with a bi-level optimization algorithm. In DARTS, it is assumed that the model can select the operation only when the network weights are properly learned so that $\boldsymbol{\alpha}$ can "make its proper decision". In the context of AutoFIS formulation, this means that we can decide whether a gate should be open or closed after the network weights are properly trained, which leads us back to the problem of fully training $2^{\mathcal{C}_{m}^{2}}$ models to make the decision. To avoid this issue, DARTS proposes to approximate the optimal value of the network weights with only one gradient descent step and train $\boldsymbol{\alpha}$ and $\boldsymbol{v}$ iteratively. 

We argue that the inaccuracy of this approximation might downgrade the performance. Therefore, instead of using bi-level optimization, we propose to optimize  $\boldsymbol{\alpha}$ and $\boldsymbol{v}$ jointly with one-level optimization. Specifically, the parameters $\boldsymbol{\alpha}$ and $\boldsymbol{v}$ are updated together with gradient descent using the training set by descending on $\boldsymbol{\alpha}$ and $\boldsymbol{v}$ based on 
%\begin{equation}
%\partial_v \mathcal{L}_{search}(v_{t-1},\alpha_{t-1})  \hspace{10pt}\text{and}\hspace{10pt} \partial_{\alpha}\mathcal{L}_{search}(v_{t-1},\alpha_{t-1}).
%\end{equation}
% \textcolor{red}{Here should we emphasis that different from two-level optimization, onelevel optimization update alpha and w with the same traning set mini-batches ?(DARTS alternatively with training and validation mini-batches)}
\begin{equation}
\partial_v \mathcal{L}_{train}(v_{t-1},\alpha_{t-1})  \hspace{10pt}\text{and}\hspace{10pt} \partial_{\alpha}\mathcal{L}_{train}(v_{t-1},\alpha_{t-1}).
\end{equation}

In this setting, $\boldsymbol{\alpha}$ and $\boldsymbol{v}$ can explore their design space freely until convergence, and $\boldsymbol{\alpha}$ is learned to serve as the contribution of individual feature interactions. In Section~\ref{sec:exp}, we would show the superiority of one-level optimization over two-level optimization.  

\textbf{Re-train Stage.}
 After the training of the \emph{search stage}, some unimportant interactions are thrown away automatically according to the architecture parameters $\boldsymbol{\alpha}^*$ in search stage. We use $\mathcal{G}_{(i,j)}$ to represent the gate status of feature interaction $\langle\textbf{\textit{e}}_{i}, \textbf{\textit{e}}_{j}\rangle$ and set $\mathcal{G}_{(i,j)}$ as $0$ when $\alpha_{(i,j)}^*=0$; otherwise, we set $\mathcal{G}_{(i,j)}$ as $1$.
In the \emph{re-train stage}, the gate status of these unimportant feature interactions are fixed to be closed permanently.

After removing these unimportant interactions, we re-train the new model with $\boldsymbol{\alpha}$ kept in the model. Specifically, we replace the feature interaction layer in Equation \ref{eqn:FM_inter1} with
%\begin{small}
\begin{equation}
    l_{fm}^{re}=\langle \boldsymbol{w},\boldsymbol{x}\rangle +\sum_{i=1}^m\sum_{j>i}^m\alpha_{(i,j)}\mathcal{G}_{(i,j)} \langle\textbf{\textit{e}}_{i}, \textbf{\textit{e}}_{j}\rangle.
    \label{eqn:FM_inter_retrain}
\end{equation}
%\end{small}
Note here $\alpha_{(i,j)}$  no longer serves as an indicator for deciding whether an interaction should be included in the model (as in \emph{search stage}). Instead, it serves as an attention unit for the architecture to learn the relative importance of the kept feature interaction. In this stage, we do not need to select the feature interactions. Therefore, all parameters are learned by Adam optimizer.

\section{EXPERIMENTS}
\label{sec:exp}
In this section, we conduct extensive offline experiments\footnote{\scriptsize Repeatable experiment code: https://github.com/zhuchenxv/AutoFIS} on two benchmark public datasets and a private dataset, as well as online A/B test, to answer the following questions:

\begin{itemize}[leftmargin = 10 pt]
    \item \textbf{RQ1}: Could we boost the performance of  factorization models with the selected interactions by AutoFIS? 
    %How about high-order feature interactions?
    % \item \textbf{RQ2}: Considering useful feature interactions selected by AutoFIS, can high-order feature interactions further improve the performance without much extra complexity?
    \item \textbf{RQ2}: Could interactions selected from simple models be transferred to the state-of-the-art models to reduce their inference time and improve prediction accuracy?
    \item \textbf{RQ3}: Are the interactions selected by AutoFIS really important and useful?
    \item \textbf{RQ4}: Can AutoFIS improve the performance of existing models in a live recommender system?
    \item \textbf{RQ5}: How do different components of our AutoFIS (e.g., BN) contribute to the performance?
\end{itemize}

\begin{table*}[t]
\centering
 \vspace{-1em}
\caption{\small Benchmark performance: "time" is the inference time for 2 million samples. "top" represents the percentage of feature interactions kept for $2^{nd}$ / $3^{rd}$ order interaction. "cost" contain the GPU time of the search and re-train stage.  "Rel. Impr." is the relative AUC improvement over FM model. Note: FFM has a lower time and cost due to its smaller embedding size limited by GPU memory constraint.}
%Note: FFM has a lower time and search cost due to its smaller embedding size to prevent overfitting; time comparison  between Avazu and Criteo makes no sense due to different hyperparameters used across such two datasets.}
% of AutoFM(2nd) over FM, AutoDeepFM(2nd) over DeepFM, AutoFM(3rd) over FM(3rd), and AutoDeepFM(3rd) over DeepFM(3rd).
% \vspace{0.5em}
 \vspace{-1em}
\label{tab:performance_overall_final}
% \vspace{0.5em}
\tiny
\resizebox{\textwidth}{!}
{
\begin{tabular}{|c|cccccc|cccccc|}
\hline
\multicolumn{1}{|c|}{\multirow{3}{*}{Model}}& \multicolumn{6}{c}{Avazu} & \multicolumn{6}{|c|}{Criteo}\\
\cline{2-13}
\multicolumn{1}{|c|}{} & \multirow{2}{*}{AUC} & \multirow{2}{*}{log loss} & \multirow{2}{*}{top} & \multirow{2}{*}{time (s)} & search + re-train
  & \multirow{2}{*}{Rel. Impr.} & \multirow{2}{*}{AUC} & \multirow{2}{*}{log loss} & \multirow{2}{*}{top} & \multirow{2}{*}{time (s)} & search + re-train  & \multirow{2}{*}{Rel. Impr.}  \\
  & & & & &\multicolumn{1}{c}{cost (min)} & & & & & &\multicolumn{1}{c}{cost (min)}  &  \\
\hline
% \multicolumn{1}{|c|}{GBDT+LR}  & 0.7721 & 0.3841 & 100\% & 0.45 & \multicolumn{1}{c}{8 + 3} & -0.92\% & 0.7871  & 0.5556 & 100\% & 0.62 & \multicolumn{1}{c}{40 + 10}  & -0.48\% \\ 
% \multicolumn{1}{|c|}{GBDT+FFM}  & 0.7835  &  0.3777 & 100\% & 2.66 &  \ \  6 + 21 & 0.54\% & 0.7988  & 0.5430 & 100\% & 1.68 &\ \  9 + 57  & 1.00\% \\ 
\multicolumn{1}{|c|}{FM}  & 0.7793 & 0.3805 & 100\% & 0.51 & 0 + 3 & 0 & 0.7909  &0.5500 & 100\% & 0.74  &\ \   0 + 11  & 0 \\ 
\multicolumn{1}{|c|}{FwFM}  & 0.7822 & 0.3784  & 100\% & 0.52 & 0 + 4 & 0.37\%  & 0.7948  &0.5475 & 100\% & 0.76 &\ \  0 + 12 & 0.49\%  \\ 
\multicolumn{1}{|c|}{AFM} &  0.7806 &0.3794 &100\% & 1.92 & \ \  0 + 14 &0.17\% & 0.7913 & 0.5517   & 100\% & 1.43 & \ \   0 + 20 &0.05\% \\
\multicolumn{1}{|c|}{FFM}  & 0.7831 & 0.3781 & 100\% &  0.24 & 0 + 6 &0.49\% & 0.7980  &0.5438 &100\% & 0.49 &\ \   0 + 39 &0.90\%  \\
\multicolumn{1}{|c|}{DeepFM}  &  0.7836 & 0.3776 & 100\% & 0.76 & 0 + 6 & 0.55\% &  0.7991 & 0.5423 & 100\% & 1.17 &\ \   0 + 16 & 1.04\%  \\
\multicolumn{1}{|c|}{GBDT+LR}  & 0.7721 & 0.3841 & 100\% & 0.45 & \multicolumn{1}{c}{8 + 3} & -0.92\% & 0.7871  & 0.5556 & 100\% & 0.62 & \multicolumn{1}{c}{40 + 10}  & -0.48\% \\ 
\multicolumn{1}{|c|}{GBDT+FFM}  & 0.7835  &  0.3777 & 100\% & 2.66 &  \ \  6 + 21 & 0.54\% & 0.7988  & 0.5430 & 100\% & 1.68 &\ \  9 + 57  & 1.00\% \\ 
% \multicolumn{1}{|c|}{IPNN} & 0.7868 & 0.3756 & 100\% & 0.91 & 6 & 0.41\% & 0.8013 &  0.5401 & 100\% & 1.26 & 19 &  0.28\% \\\hline
\multicolumn{1}{|c|}{AutoFM(2nd)}  & 0.7831*  & 0.3778* & 29\% & \textbf{0.23} & 4 + 2 &  0.49\% &  0.7974* & 0.5446* & 51\% & \textbf{0.48}  & 14 + 9\ \   & 0.82\% \\
\multicolumn{1}{|c|}{AutoDeepFM(2nd)}  &  \textbf{0.7852*} & \textbf{0.3765*} & \textbf{24\%} &  0.48 & 7 + 4 & 0.76\%  & \textbf{0.8009*}  & \textbf{0.5404*} & \textbf{28\%} & 0.69 & 22 + 11 & 1.26\% \\ \hline\hline
\multicolumn{1}{|c|}{FM(3rd)}  & 0.7843 & 0.3772 &  100\% & 5.70 &\ \   0 + 21  & 0.64\% & 0.7965  &0.5457 &  100\% & 8.21 &\ \   0 + 72 & 0.71\% \\
\multicolumn{1}{|c|}{DeepFM(3rd)}  & 0.7854 & 0.3765 & 100\% & 5.97 & \ \  0 + 23 & 0.78\% & 0.7999  &0.5418 &  100\% & 13.07 &\ \  \ \   0 + 125 & 1.14\% \\\hline

\multicolumn{1}{|c|}{AutoFM(3rd)}  &  0.7860* & 0.3762* & \textbf{25\% / 2\%} & \textbf{0.33} & 22 + 5\ \   &  0.86\% & 0.7983*  & 0.5436* & 35\% / 1\% & \textbf{0.63} & 75 + 15 & 0.94\%  \\
\multicolumn{1}{|c|}{AutoDeepFM(3rd)}  & \textbf{0.7870*}  & \textbf{0.3756*} & 21\% / 10\% & 0.94 & \;24 + 10 & 0.99\%  & \textbf{0.8010*}  & \textbf{0.5404*} & \textbf{13\% / 2\%} &  0.86 &  128 + 17\ \   & 1.28\%  \\
\hline
\end{tabular}

 }
 \begin{tablenotes}
		 \item \scriptsize  $*$ denotes statistically significant improvement (measured by t-test with p-value$<$0.005) over  baselines with same order. AutoFM compares with FM and AutoDeepFM compares with all baselines.
	\end{tablenotes}
 \vspace{-2.5em}
\end{table*}

\subsection{Datasets}
Experiments are conducted for the following two public datasets (Avazu and Criteo) and one private dataset:

\textbf{Avazu\footnote{{\scriptsize http://www.kaggle.com/c/avazu-ctr-prediction}}}: Avazu was released in the CTR prediction contest on Kaggle. $80\%$ of randomly shuffled data is allotted to training and validation with $20\%$ for testing. Categories with less than 20 times of appearance are removed for dimensionality reduction.

\textbf{Criteo\footnote{{\scriptsize http://labs.criteo.com/downloads/download-terabyte-click-logs/}}}: Criteo contains one month of click logs with billions of data samples. We select "data 6-12" as training and validation set while selecting "day-13" for evaluation. To counter label imbalance, negative down-sampling is applied to keep the positive ratio roughly at $50\%$. 13 numerical fields are converted into one-hot features through bucketing, where the features in a certain field appearing less than 20 times are set as a dummy feature "other".

\textbf{Private}: Private dataset is collected from a game recommendation scenario in Huawei App Store. The dataset contains app features (e.g., ID, category), user features (e.g., user's behavior history) and context features.
Statistics of all the datasets are summarized in Table~\ref{tab:dataset}.
\begin{table}[h]
\vspace{-1em}
\caption{\small Dataset Statistics}
\vspace{-1em}
\scriptsize
\label{tab:dataset}
\centering
\resizebox{0.38\textwidth}{!}{
\begin{tabular}{|l|c|c|c|c|}
\hline
Dataset & \#instances & \#dimension & \#fields & pos ratio \\ \hline
%Criteo & $1\times 10^{8}$ & $1\times 10^{6}$ & 39 & 0.50 \\
Avazu & $4\times 10^{7}$ & $6\times 10^{5}$ & 24 & 0.17 \\
Criteo & $1 \times 10^{8}$ & $1\times 10^{6}$ & 39 & 0.50 \\
Private & $2\times 10^{8}$ & $3\times 10^{5}$ & 29 & 0.02 \\\hline
\end{tabular}
}
\vspace{-3em}
\end{table}

\subsection{Experimental Settings}
% \subsubsection{Evaluation Metrics}
% The common evaluation metrics for CTR prediction are \textbf{AUC} (Area Under ROC) and \textbf{Log loss} (cross-entropy).

% \textbf{AUC}: Area Under ROC curve is a widely used metric in evaluating classification problems. Besides, some work validates AUC as a good measurement in CTR estimation~\cite{graepel2010web}.

% \textbf{Log loss}: Log loss measures the performance of a classification model where the prediction is a probability value between $0$ and $1$.

\subsubsection{Baselines and Evaluation Metrics}
 We apply AutoFIS to FM~\cite{fm} and DeepFM~\cite{deepfm} models to show its effectiveness (denoted as \textbf{AutoFM} and \textbf{AutoDeepFM}, respectively). We compare it with GBDT-based methods (GBDT+LR~\cite{gbdtlr}, GBDT+FFM~\cite{juan2014idiots}) and Factorization Machine models (AFM~\cite{afm}, FwFM~\cite{fwfm}, FFM~\cite{ffm}, IPNN~\cite{pin}). Due to its huge computational costs and the unavailability of the source code, we do not compare our models with AutoCross~\cite{autocross}.
 
 The common evaluation metrics for CTR prediction are \textbf{AUC} (Area Under ROC) and \textbf{Log loss} (cross-entropy).
%  The baseline methods are detailed as follows:
 
%  \begin{itemize}[leftmargin = 10 pt]
%     \item \textbf{GBDT+LR/FFM}: Gradient Boosting Decision Tree (GBDT) transforms feature and then LR (Logistic Regression)/FFM (Field-aware Factorization Machine) classifies the instances based on the transformed feature interactions.
%     %\item \textbf{GBDT+FFM}: Gradient Boosting Decision Tree (GBDT) transforms feature and use Field-aware Factorization Machine to classify the transformed feature interactions .
%     \item \textbf{FM}: Factorization Machine (FM) is a benchmark factorization model, which considers the second-order feature interactions.
%     \item \textbf{FFM}: Field-aware Factorization Machine (FFM)~\cite{ffm} is a variant of FM, which considers the conception of field. The feature embedding is not only related to the feature but also to the interacted fields.
%     \item \textbf{FwFM}: Field-weighted Factorization Machine (FwFM)~\cite{fwfm} models different second-order feature interactions by different weights via MLP.
%     \item \textbf{AFM}: Attentional Factorization Machine (AFM)~\cite{afm} learns the importance of feature interactions via an attention network.
%     \item \textbf{DeepFM}: Deep Factorization Machine (DeepFM) is a parallel combination of FM model and MLP layer.
%     \item \textbf{IPNN}: Inner Product Neural Network (IPNN) is a sequential combination of FM model and MLP layer.
% \end{itemize}

\subsubsection{Parameter Settings}

To enable any one to reproduce the experimental results, we have attached all the hyper-parameters for each model in the supplementary material.
% Table~\ref{tab:parameter} summarizes the parameters of all models.
% For Criteo and Avazu datasets, the parameters of baseline models are set following~\cite{pin}. For AutoFM and AutoDeepFM we use the same hyper-parameters as the base models (i.e., FM and DeepFM accordingly) except for extra ones in AutoFIS.

\subsubsection{Implementation Details}

Selecting $2^{nd}$-order feature interactions for AutoFM and AutoDeepFM, in the search stage, we first train the model with $\boldsymbol{\alpha}$ and $\boldsymbol{v}$ jointly on all the training data. Then we remove those useless interactions and re-train our model.

To implement AutoFM and AutoDeepFM for $3^{rd}$-order feature interaction selection, we reuse the selected $2^{nd}$-order interactions in Equation \ref{eqn:FM_inter_retrain} and enumerate all the $3^{rd}$-order feature interactions in the search stage to learn their importance.
%We try to reduce the number of the selected $2^{nd}$ and $3^{rd}$-order interactions simultaneously. 
Finally, we re-train our model with the selected $2^{nd}$- and $3^{rd}$-order interactions.

Note that in the search stage, the architecture parameters $\boldsymbol{\alpha}$ are optimized by GRDA optimizer and other parameters $\boldsymbol{v}$ are optimized by Adam optimizer. In the re-train stage, all parameters are optimized by Adam optimizer.

\begin{table}[h]
\vspace{-1em}
   \caption{\small Performance in Private Dataset. "Rel. Impr." is the relative AUC improvement over FM model.}
\vspace{-1em}
\label{tab:performance_private}
\small
% \resizebox{0.35\textwidth}{!}
% {
\begin{center}
{
\resizebox{0.35\textwidth}{!}{
\begin{tabular}{|c|cccc|}
\hline
% \multicolumn{1}{|c|}{\multirow{2}{*}{Model}}&  \multicolumn{4}{|c|}{Private}\\
% \cline{2-4}
% \multicolumn{1}{|c|}{} &  AUC & log loss & top & Rel. Impr. \\
% \hline
\multicolumn{1}{|c|}{Model} & AUC &log loss & top & ReI. Impr\\
\hline
\multicolumn{1}{|c|}{FM}  &  0.8880 & 0.08881 & 100\%&0 \\ 
\multicolumn{1}{|c|}{FwFM}  & 0.8897& 0.08826 & 100\% &0.19\% \\ 
\multicolumn{1}{|c|}{AFM} & 0.8915 & 0.08772 & 100\%&0.39\% \\
\multicolumn{1}{|c|}{FFM}  & 0.8921 & 0.08816 & 100\%&0.46\% \\
\multicolumn{1}{|c|}{DeepFM}  & 0.8948 & 0.08735 & 100\%&0.77\%  \\\hline
\multicolumn{1}{|c|}{AutoFM(2nd)}  & 0.8944*  &   0.08665* & 37\% &0.72\% \\
\multicolumn{1}{|c|}{AutoDeepFM(2nd)}  &\textbf{0.8979*} &  \textbf{0.08560*} & \textbf{15\%} &1.11\% \\ 
\hline
\end{tabular}
}
}
\end{center}
 \begin{tablenotes}
		 \item \scriptsize \qquad \quad  $*$ denotes statistically significant improvement (measured by t-test with p-value$<$0.005).
		 \item \scriptsize \qquad \quad AutoFM compares with FM and AutoDeepFM compares with all baselines.
	\end{tablenotes}
\vspace{-2em}
\end{table}

\subsection{Feature Interaction Selection by AutoFIS (RQ1)}
Table~\ref{tab:performance_overall_final} summarizes the performance of AutoFM and AutoDeepFM by automatically selecting $2^{nd}$- and $3^{rd}$-order important interactions on Avazu and Criteo datasets and Table~\ref{tab:performance_private} reports their performance on Private dataset. We can observe: 
\begin{enumerate}[leftmargin = 10 pt]
% \item FwFM and AFM outperform FM consistently, which indicates that learning a weight for each feature interaction in can boost the performance. In ablation study, we will discuss why the weights in FwFM can not be used for interaction selections. 
\item For Avazu dataset, 71\% of the $2^{nd}$-order interactions can be removed for FM and 76\% for DeepFM. Removing those useless interactions can not only make the model faster at inference time: the inference time of AutoFM(2nd) and AutoDeepFM(2nd) is apparently less than FM and DeepFM; but also significantly increase the prediction accuracy: the relative performance improvement of AutoFM(2nd) over FM is 0.49\% and that of AutoDeepFM(2nd) over DeepFM is 0.20\% in terms of AUC. Similar improvement can also be drawn from the other datasets. 
\item For high-order feature interaction selection, only 2\% -- 10\% of all the $3^{rd}$-order feature interactions need to be included in the model.
% with similar inference time with the original FM and deepFM models correspondingly. 
The inference time of AutoFM(3rd) and AutoDeepFM(3rd) is much less than that of FM(3rd) and DeepFM(3rd) (which is comparable to FM and DeepFM). Meanwhile, the accuracy is significantly improved by removing unimportant $3^{rd}$-order feature interactions, i.e., the relative performance improvement of AutoFM(3rd) over FM(3rd) is 0.22\% and that of AutoDeepFM(3rd) over DeepFM(3rd) is 0.20\% in terms of AUC on Avazu. Observations on Criteo are similar.   
\item All such performance boost could be achieved with marginal time cost (for example, it takes 24 minutes and 128 minutes for AutoDeepFM(3rd) to search important $2^{nd}$- and $3^{rd}$-order feature interactions in Avazu and Criteo with a single GPU card). The same result might take the human engineers many hours or days to achieve by identifying such important feature interactions manually. 

\end{enumerate}
Note that directly enumerating the $3^{rd}$-order feature interactions in FM and DeepFM enlarges the inference time about 7 to 12 times, which is unacceptable in industrial applications.

% \ruiming{IPNN? Significant test?}

\subsection{Transferability of the  Selected Feature Interactions (RQ2)}

\begin{table}[h]
\centering
% \scriptsize
\vspace{-1em}
\caption{\small Performance of transferring interactions selected by AutoFM to IPNN. 
AutoIPNN(2nd) indicates IPNN with $2^{nd}$-order interactions selected by AutoFM(2nd) and AutoIPNN(3rd) indicates IPNN with $2^{nd}$- and $3^{rd}$-order interactions selected by AutoFM(3rd).}     
% $^*$: with the selected $3^{rd}$-order interactions, we can use fewer $2^{nd}$-order interactions, resulting in less computation cost.
\small
\label{tab:performance_ipnn}
\vspace{-1em}
\centering
\resizebox{0.45\textwidth}{!}{
\begin{tabular}{l|lll|lll|}
\hline
\multicolumn{1}{|c|}{\multirow{2}{*}{Model}}& \multicolumn{3}{c|}{Avazu} & \multicolumn{3}{c|}{Criteo}\\
\cline{2-7}
\multicolumn{1}{|c|}{} & AUC & log loss  & time(s) & AUC & log loss &  time(s)\\
\hline
\multicolumn{1}{|c|}{IPNN}  & 0.7868 & 0.3756 & 0.91 & 0.8013  &0.5401  & 1.26 \\ 
\multicolumn{1}{|c|}{AutoIPNN(2nd)}  &  0.7869 & 0.3755  & 0.58 & 0.8015  & 0.5399 & 0.76  \\
\multicolumn{1}{|c|}{AutoIPNN(3rd)} & \textbf{0.7885*} & \textbf{0.3746*} & 0.71 & \textbf{0.8019*}  & \textbf{0.5392*} & 0.86 \\
\hline
\end{tabular}
}
 \begin{tablenotes}
		 \item \scriptsize $*$ denotes statistically significant improvement (measured by t-test with p-value$<$0.005).
	\end{tablenotes}
\vspace{-1em}
\end{table}
In this subsection, we investigate whether the feature interactions learned by AutoFM (which is a simple model) could be transferred to the state-of-the-art models such as IPNN to boost their performance. As shown in Table~\ref{tab:performance_ipnn}, using $2^{nd}$-order feature interactions selected by AutoFM (namely AutoIPNN(2nd)) achieves comparable performance to IPNN, with around 30\% and 50\% of all the interactions in Avazu and Criteo. Moreover, the performance is significantly improved by using both $2^{nd}$- and $3^{rd}$-order feature interactions (namely AutoIPNN(3rd)) selected by AutoFM. Both evidences verify the transferability of the selected feature interactions in AutoFM. 
% In this subsection, we investigate whether the feature interactions learned by AutoFM (which is a simple model) could be transferred to the state-of-the-art models such as IPNN to boost their performance. As shown in Table~\ref{tab:performance_ipnn}, using $2^{nd}$-order feature interactions selected by AutoFM (namely IPNN(2)) achieves comparable performance to IPNN, with around 30\% and 50\% of all the interactions in Avazu and Criteo. Moreover, the performance is significantly improved by using both $2^{nd}$- and $3^{rd}$-order feature interactions (namely IPNN(3)) selected by AutoFM. Both evidences verify the transferability of the selected feature interactions in AutoFM. 

% \ruiming{Significant test?}

\subsection{The Effectiveness of Feature Interaction Selected by AutoFIS (RQ3)}
In this subsection, we will discuss the effectiveness of feature interaction selected by AutoFIS. We conduct experiments on real data and synthetic data to analyze it.

\subsubsection{The Effectiveness of selected feature interaction on Real Data}
We define \textbf{$\mbox{statistics}\_\mbox{AUC}$} to represent the importance of a feature interaction to the final prediction. For a given interaction, we construct a predictor only considering this interaction where the prediction of a test instance is the statistical CTR ($\#downloads / \#impressions$) of specified feature interaction in the training set. Then the AUC of this predictor is $\mbox{statistics}\_\mbox{AUC}$ with respect to this given feature interaction. Higher $\mbox{statistics}\_\mbox{AUC}$ indicates a more important role of this feature interaction in prediction.
Then we visualize the relationship between $\mbox{statistics}\_\mbox{AUC}$ and $\alpha$ value.

% figure
\begin{figure}[h]
    \centering
    \vspace{-1em}
    \includegraphics[width=0.38\textwidth]{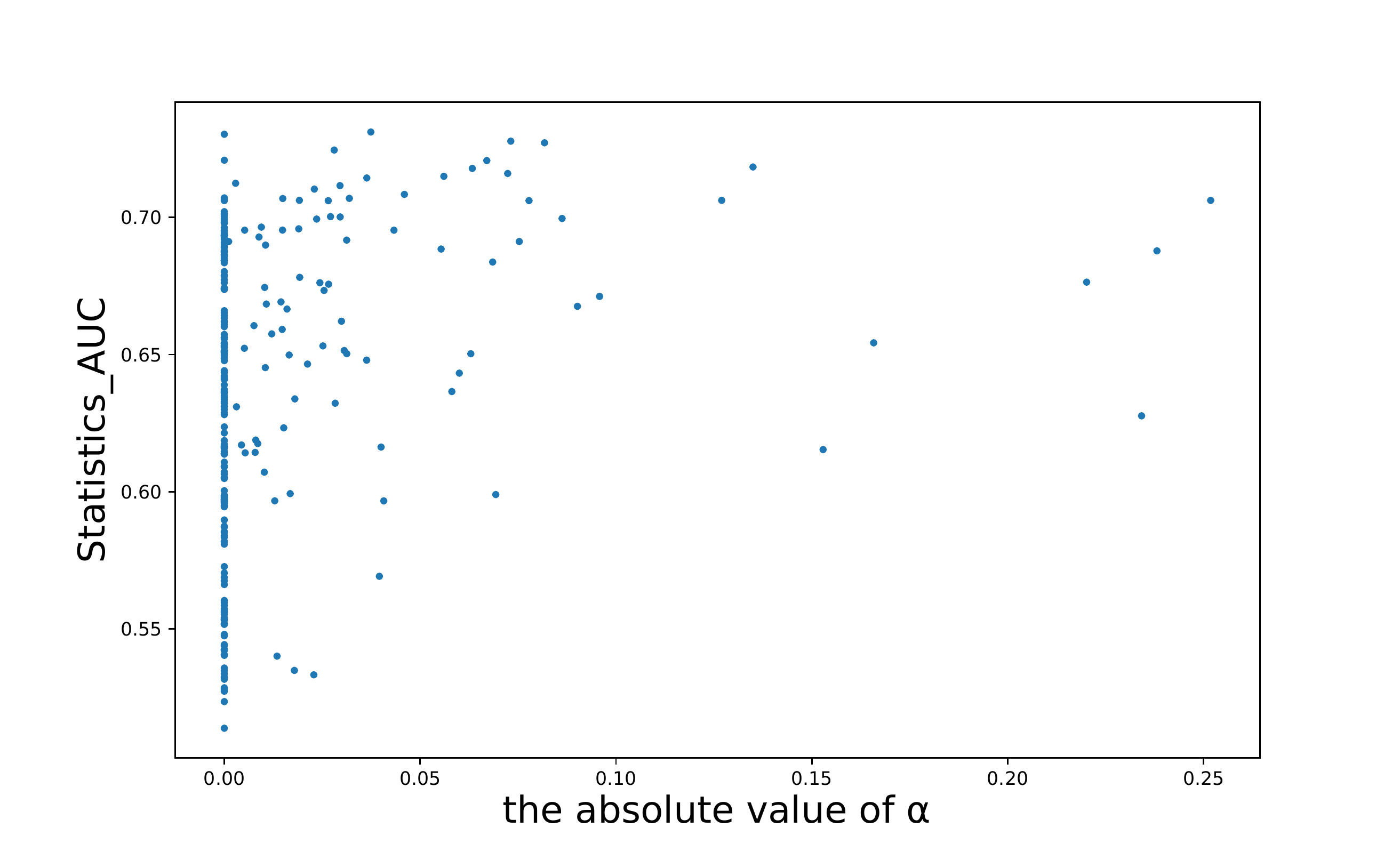}
    % \includegraphics[width=7cm,height=5cm]{picture/real_data_auc_alpha.pdf}
    % \vspace{0em}
    \vspace{-1em}
        \caption{\small Relationship between $\mbox{statistics}\_\mbox{AUC}$ and $\alpha$ value for each second-order interaction}
    % \ruiming{characters in the figure}}
    \vspace{-1em}
    \label{fig:auc_alpha}
\end{figure}

As shown in Figure~\ref{fig:auc_alpha}, we can find that most of the feature interactions selected by our model (with high absolute $\alpha$ value) have high $\mbox{statistics}\_\mbox{AUC}$, but not all feature interactions with high $\mbox{statistics}\_\mbox{AUC}$ are selected. That is because the information in these interactions may also exist in other interactions which are selected by our model.

\begin{table}[h]
    \small
    \centering
    % \vspace{-1em}
    \caption{\small Performance comparison between the model with interactions selected by our model and by $\mbox{statistics}\_\mbox{AUC}$ on Avazu Dataset}
    \vspace{-1em}
    \begin{tabular}{|c|c|c|}
    \hline
        Model & AUC & log loss  \\\hline
       
        Selected by $\mbox{statistics}\_\mbox{AUC}$ & 0.7804 & 0.3794    \\\hline
         Selected by AutoFM &  0.7831 & 0.3778 \\\hline
        \end{tabular}
    \label{tab:auc_top}
    \vspace{-1em}
\end{table}

To evaluate the effectiveness of the selected interactions by our model, we also select the top-$N$ ($N$ is the number of second-order feature interactions selected by our model) interactions based on $\mbox{statistics}\_\mbox{AUC}$ and re-train the model with these interactions. As shown in Table~\ref{tab:auc_top}, the performance of our model is much better than the model with selected interactions by $\mbox{statistics}\_\mbox{AUC}$ with same computational cost.

\subsubsection{The Effectiveness of selected feature interaction on Synthetic Data}

In this section, we conduct a synthetic experiment to validate the effectiveness of selected feature interaction. 

This synthetic dataset is generated from an incomplete poly-2 function, where the bi-linear terms are analogous to interactions
between categories. Based on this dataset, we investigate (i) whether our model could find the important interactions (ii) the performance of our model compared with other factorization machine models.

The input $x$ of this dataset is randomly sampled from $N$ categories of $m$ fields.
%, where each field size $N_i$ is randomly selected. 
The output $y$ is binary labeled depending on the sum of linear terms and parts of bi-linear terms. 
%The bi-linear set $C$ is randomly selected.

\begin{equation}
    y = \delta{(\sum_{i=1}^{m}w_{i}x_{i}+\sum_{i,j\in C}v_{i,j}x_ix_j+b+\epsilon)}
\end{equation}

\begin{equation}
    \delta(z)=\left\{
    \begin{aligned}
&1, \quad if\; z\geq threshold\\
&0, \quad otherwise
\end{aligned}
\right. 
\end{equation}
The data distribution $p(x)$, selected bi-linear term sets $C$ and $\boldsymbol{w}, \boldsymbol{v}, b$ are randomly sampled and fixed. The data pairs are $i.i.d.$ sampled to build the training and test datasets. We also add a small random noise $\epsilon$ to the sampled data.
We use FM and our model to fit the synthetic data. We use AUC to evaluate these models on the test dataset.

% \begin{figure}[t]
%     \centering
%     \includegraphics[width=0.48\textwidth]{picture/combination.png}
%     %\vspace{-2.5em}
    
%     \caption{Training results of the synthetic experiments. Note: Plot (a) verifies AutoFM as a better model. Plot (b) verifies AutoFM could find more important interactions.}
%     % \vspace{-1em}
%     \label{fig:synthetic_exp}
% \end{figure}

\begin{figure}[h]
    \centering
    \vspace{-1em}
    \includegraphics[width=0.38\textwidth]{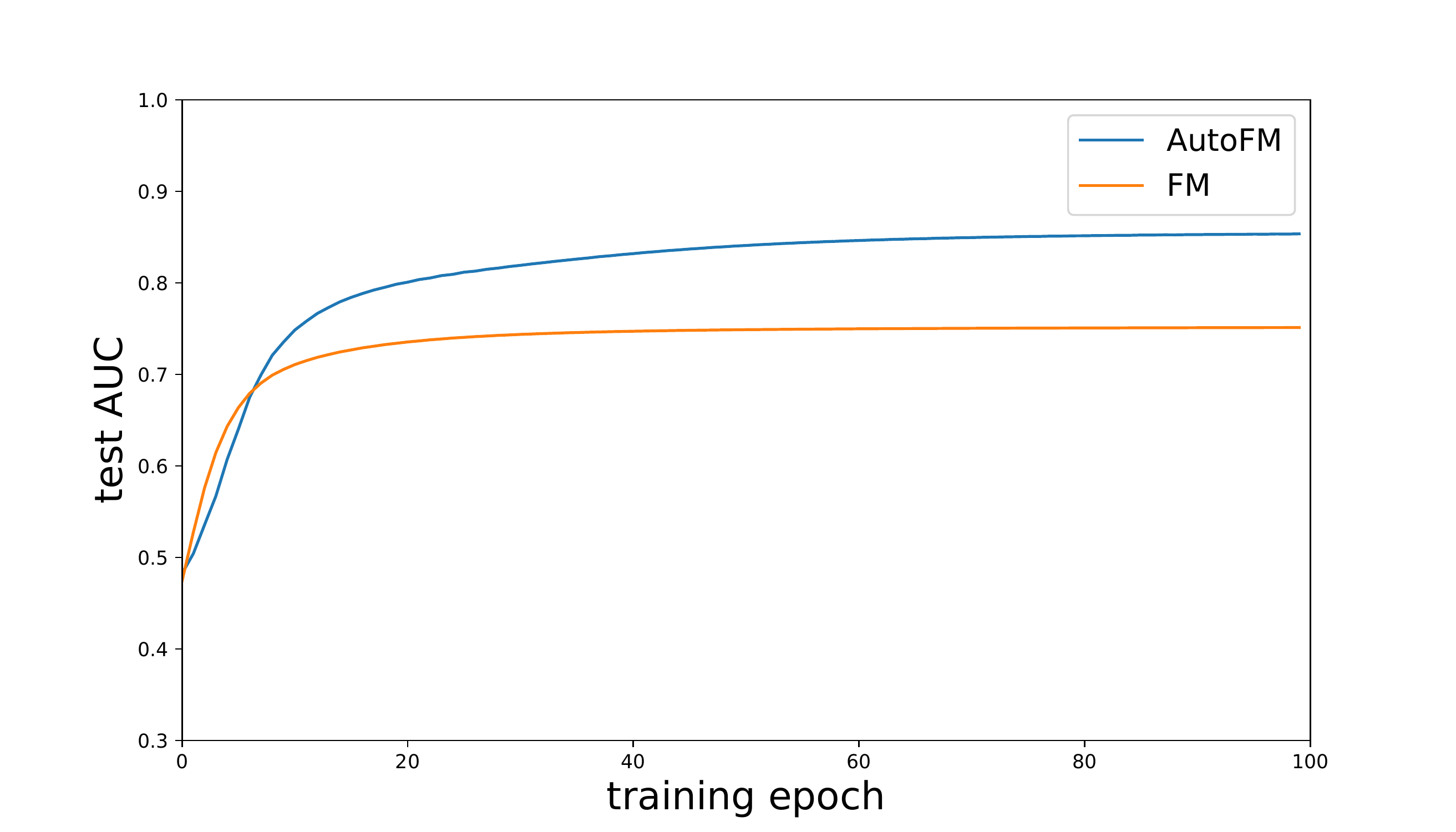}
    \vspace{-1em}
    
    \caption{\small Training results of the synthetic experiments to verify AutoFM as a better model.}
    \vspace{-1em}
    \label{fig:synthetic_exp_a}
\end{figure}

\begin{figure}[h]
    \centering
    \vspace{-0.5em}
    \includegraphics[width=0.38\textwidth]{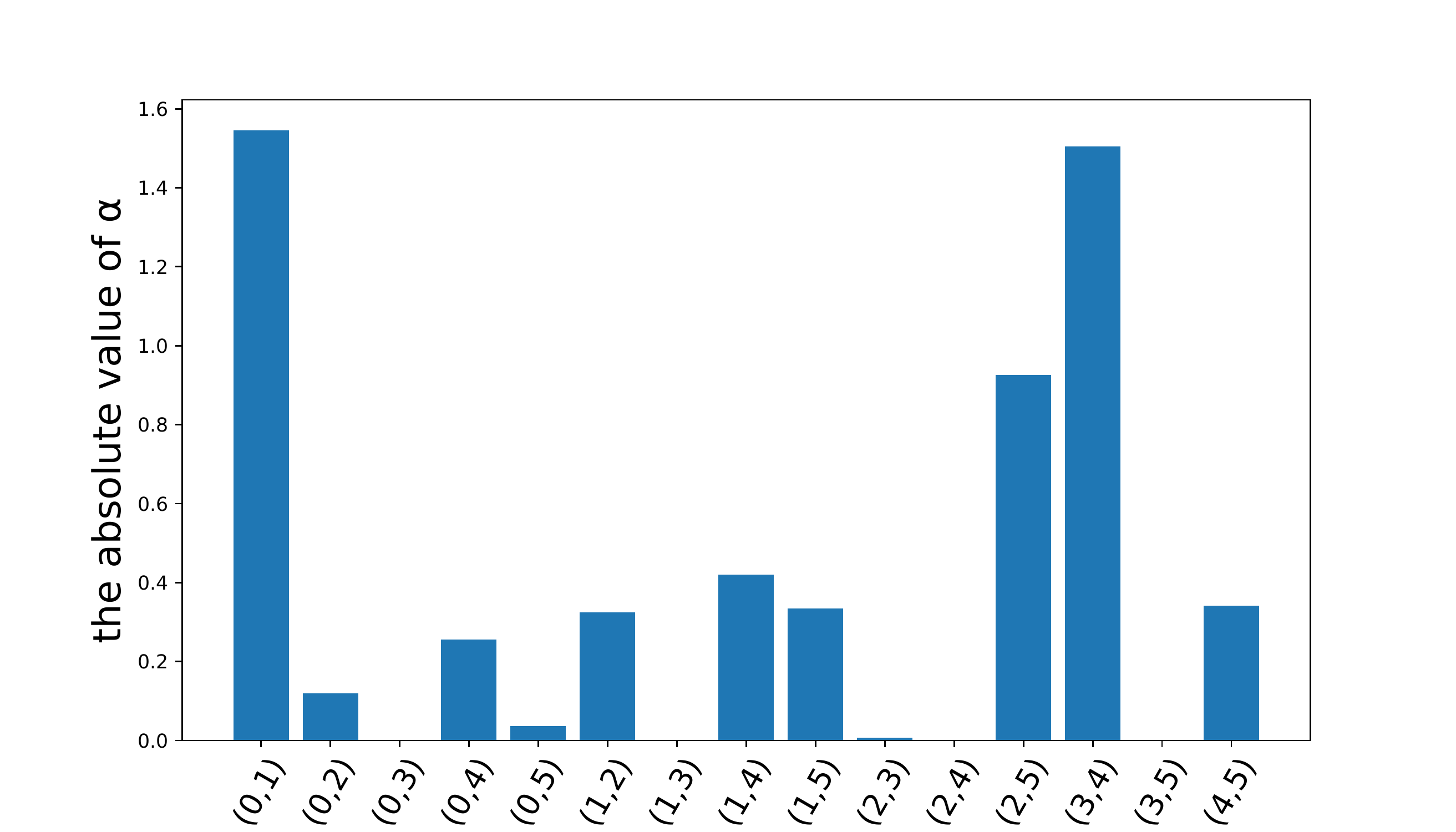}
    % \vspace{-2.5em}
    \vspace{-1em}
    \caption{\small Training results of the synthetic experiments to verify AutoFM could find more important interactions.} %\ruiming{\alpha}}
    \vspace{-1.5em}
    \label{fig:synthetic_exp_b}
\end{figure}

We choose $m= 6, N = 60$ to test the effectiveness of our model. Selected bi-linear term sets $C$ is randomly initialized as $C = \{(x_0, x_1), (x_2, x_5), (x_3, x_4)\}$. Figure~\ref{fig:synthetic_exp_a} presents the performance comparison between our model and FM, which demonstrates the superiority of our model. As shown in Figure~\ref{fig:synthetic_exp_b}, our model could extract the important interactions precisely. The interactions in $C$ have the highest $\alpha$ and some unimportant interactions (with $\alpha$ value 0) have been removed.

\subsection{Deployment \& Online Experiments (RQ4)}
Online experiments were conducted in the recommender system of Huawei App Store to verify the superior performance of AutoDeepFM. Huawei App Store has hundreds of millions of daily active users which generates hundreds of billions of user log events everyday in the form of implicit feedback such as browsing, clicking and downloading apps. In online serving system, hundreds of candidate apps that are most likely to be downloaded by the users are selected by a model from the universal app pool. These candidate apps are then ranked by a fine-tuned ranking model (such as DeepFM, AutoDeepFM) before presenting to users. To guarantee user experience, the overall latency of the above-mentioned candidate selection and ranking is required to be within a few milliseconds. To deploy AutoDeepFM, we utilize a three-node cluster, where each node is with 48 core Intel Xeon CPU E5-2670 (2.30GHZ), 400GB RAM and as well as 2 NVIDIA TESLA V100 GPU cards.

%\textcolor{red}{Ruiming: The OS of the machines is not mentioned, as it is developed by Huawei and I could not find reference.}\weinan{ok.}

Specifically, a ten-day AB test is conducted in a game recommendation scenario in the App Store. Our baseline in online experiments is DeepFM, which is a strong baseline due to its extraordinary accuracy and high efficiency which has been deployed in the commercial system for a long time. 

For the control group, 5\% of users are randomly selected and presented with recommendation generated by DeepFM. DeepFM is chosen as a strong baseline due to its extraordinary accuracy and high efficiency, which has been deployed in our commercial system for a long time. For the experimental group, 5\% of users are presented with recommendation generated by AutoDeepFM.
%~\footnote{Compared with DeepFM, AutoDeepFM only uses 20\% $2^{nd}$-order interactions.}.

% CTR($\frac{\#downloads}{\#impressions}$) and CVR($\frac{\#downloads}{\#users}$) are used to compare their online performance. 

\begin{figure}[t]
    \centering
    \vspace{-1em}
    \includegraphics[width=0.43\textwidth]{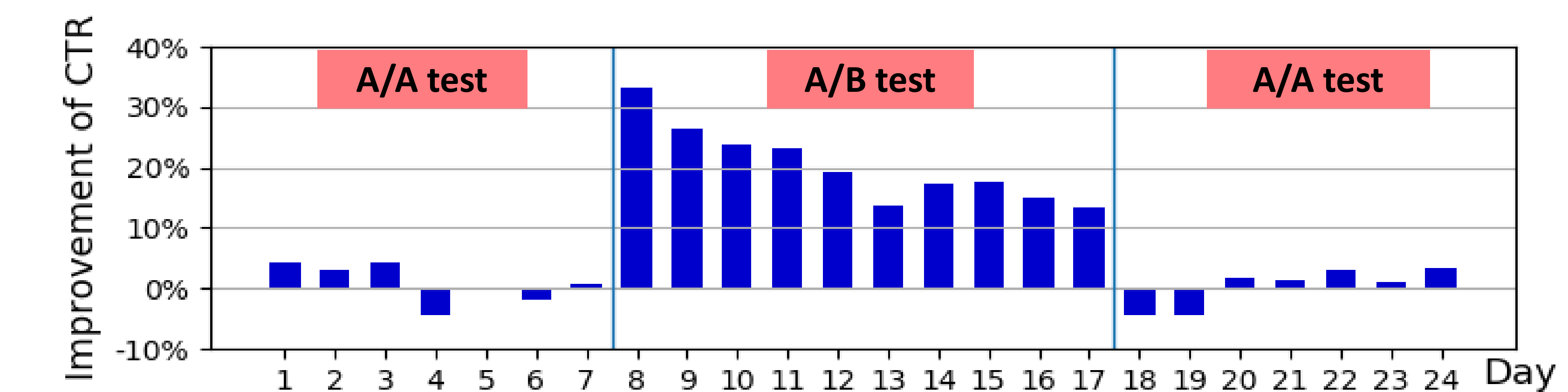}
    % \vspace{-2.5em}
    \vspace{-1em}
    \caption{\small Online experimental results of CTR.}
    \vspace{-0.5em}
    \label{fig:online_ctr}
\end{figure}

\begin{figure}[t]
    \centering
    \vspace{-0.5em}
    \includegraphics[width=0.43\textwidth]{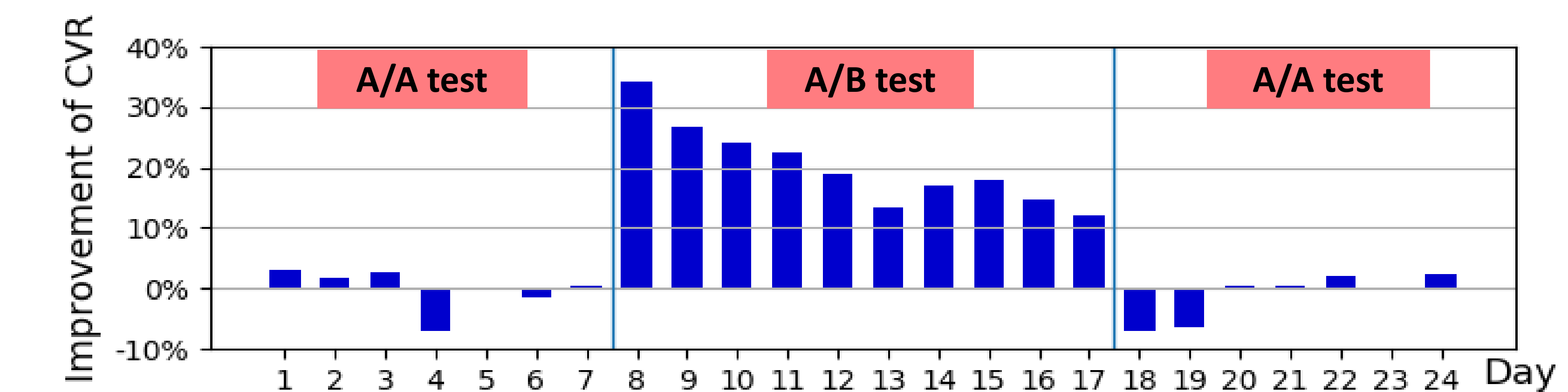}
    % \vspace{-2.5em}
    \vspace{-1em}
    \caption{\small Online experimental results of CVR.}
    \vspace{-1.5em}
    \label{fig:online_cvr}
\end{figure}

Figure~\ref{fig:online_ctr} and Figure~\ref{fig:online_cvr} show the improvement of the experimental group over the control group with CTR ($\#downloads / \#impressions$) and CVR ($\#downloads/\#users$) respectively. We can see that the system is rather stable where both CTR and CVR fluctuated within $8\%$ during the A/A testing. Our AutoDeepFM model is launched to the live system on Day 8. From Day 8, we observe a significant improvement over the baseline model with respect to both CTR and CVR. The average improvement of CTR is \textbf{20.3\%} and the average improvement of CVR is \textbf{20.1\%} over the ten days of A/B test. These results demonstrate the magnificent effectiveness of our proposed model. From Day 18, we conduct again A/A test to replace our AutoDeepFM model with the baseline model in the experimental group. We observe a sharp drop in the performance of the experimental group, which once more verifies that the improvement of online performance in the experimental group is indeed introduced by our proposed model.

% \begin{table}[h]
%     \centering
%     \caption{Results from Online A/B testing}
%     \begin{tabular}{|c|c|c|}
%     \hline
%         Model & CTR gain & CVR gain  \\\hline
%         DeepFM &  0\% & 0\% \\\hline
%         AutoDeepFM & 20.3\% & 20.1\% \\\hline
%         \end{tabular}
%     \label{tab:online}
%     \vspace{0.1em}
% \end{table}

% As shown in Table~\ref{tab:online}, compared with DeepFM, AutoDeepFM has improved CTR ($\#downloads / \#impressions$) by \textbf{20.3\%} and CVR ($\#downloads/\#users$) by \textbf{20.1\%}. Currently, AutoDeepFM has been deployed online, contributing a significant business revenue growth.

% AutoDeepFM has improved CTR($\frac{\#downloads}{\#impressions}$) by \textbf{20.3\%} and CVR($\frac{\#downloads}{\#users}$) by \textbf{20.1\%} compared with DeepFM. Currently, AutoDeepFM has been deployed online, contributing a significant business revenue growth.

\subsection{Ablation Study (RQ5)}

\subsubsection{Stability of $\boldsymbol{\alpha}$ estimation across different seeds}
In this part, we conduct experiments to check whether the trained value of $\boldsymbol{\alpha}$ is stable across different random initializations. A stable estimation of $\boldsymbol{\alpha}$ means that the model's decision on which interaction is important is not affected by the random seed. We run the \textit{search stage} of AutoFM with different seeds on Avazu. The Pearson correlation of $\alpha$ estimated from different seeds is around 0.86, this validates that the estimation of  $\boldsymbol{\alpha}$ is stable. Without the use of BN for the feature interaction (which is essentially FwFM model), this Pearson correlation drop to around 0.65. 
% that different seeds can almost produce the same selection result. 

% We also investigated the performance of using different number of feature interactions selected by AutoFIS and random selection. The performance of AutoFIS dominates. The results are attached in the Supplementary materials. 
%\subsubsection{Feature interaction selection by AutoFIS verse with Random Selection}

\subsubsection{Effectiveness of components in AutoFIS}

\begin{table}[h]
    \centering
    % \tiny
    \small
    \vspace{-1em}
    \caption{\small Different Variants.}
    \vspace{-1em}
    \resizebox{0.75\columnwidth}{!}{
    \begin{tabular}{|c|cc|cc|}
    \hline
    \multicolumn{1}{|c|}{\multirow{2}{*}{Variants}} & \multicolumn{2}{c|}{\emph{search stage}} & \multicolumn{2}{c|}{\emph{re-train stage}} \\
    \cline{2-5}
    & AutoFIS & Random & BN & $\alpha$  \\\hline
        AutoFM    &  $\surd$ & $\times$ & $\surd$ & $ \surd$ \\\hline
        AutoFM-BN & $\surd$ & $\times$ & $\times$ & $ \surd$ \\\hline
        AutoFM-BN-$\alpha$ & $\surd$ & $\times$ & $\times$ & $\times$ \\\hline
        Random+FM & $\times$  & $\surd$ & $\times$ & $\times$ \\\hline
        %FM &  $\times$  & $\times$ & $\times$ & $\times$ \\\hline
    \end{tabular}}
    \label{tab:variants}
    \vspace{-2em}
\end{table}
\begin{table}[h]
    \centering
    \small
    % \tiny
    % \vspace{-0.5em}
    \caption{\small Performance comparison of different feature interaction selection strategies. $^*$: with fewer interactions, FM may have better performance.}
    \vspace{-1em}
    \resizebox{0.75\columnwidth}{!}{
   \begin{tabular}{c|cc|cc|}
\hline
\multicolumn{1}{|c|}{\multirow{2}{*}{Model}}& \multicolumn{2}{c|}{Avazu} & \multicolumn{2}{c|}{Criteo}\\
\cline{2-5}
\multicolumn{1}{|c|}{} & AUC & log loss  & AUC & log loss  \\
\hline
\multicolumn{1}{|c|}{FM}  &  0.7793 & 0.3805 & 0.7909  & 0.5500     \\ 
\hline
\multicolumn{1}{|c|}{AutoFM}  & \textbf{0.7831}  &\textbf{0.3778} &     \textbf{0.7974} & \textbf{0.5446} \\ \hline
\multicolumn{1}{|c|}{AutoFM-BN}  & 0.7824   & 0.3783 & 0.7971  &    0.5450\\ 
\hline
\multicolumn{1}{|c|}{AutoFM-BN-$\alpha$}  &  0.7811 & 0.3793 & 0.7946  &  0.5481 \\ 
\hline
\multicolumn{1}{|c|}{Random+FM}  &  0.7781 & 0.3809 & $0.7940^*$  & 0.5486    \\ 
\hline
\end{tabular}
}
\vspace{-1em}
\label{tab:top_comparison}
\end{table}

To validate the effectiveness of individual components in AutoFIS, we propose several variants, which are enumerated in Table~\ref{tab:variants}. Recall that AutoFIS has two stages: \emph{search stage} and \emph{re-train stage}.
To verify the effectiveness of the \emph{search stage} of AutoFIS, we compare it with "Random" strategy, which selects feature interactions randomly. Similarly, in the \emph{re-train stage}, we validate the advantages of BN and $\alpha$. The relationship between different components in the two stages is presented in Table~\ref{tab:variants}. The performance of such variants presented in Table~\ref{tab:top_comparison}. 
Note that for "Random" strategy, we choose the same number of interactions with AutoFM, and we try ten different "Random" strategies and average the results.
We can get several conclusions:
\begin{enumerate}[leftmargin = 15 pt]
\item Comparing  AutoFM-BN-$\alpha$ with Random+FM, we can see that selection by AutoFIS can always achieve better performance than Random selection with same number of interactions. It demonstrates that important interactions are identified by AutoFIS in the \emph{search stage}.
% in the \emph{search stage}, selecting feature interactions by AutoFIS achieves better performance than selecting feature interactions from random weights. It demonstrates that AutoFIS selects useful feature interactions in the \emph{search stage}. 
% As it shows, the "Random" strategy selects different feature interactions via different random seeds, which are presented as "Random+FM$_1$" and "Random+FM$_2$ in Table~\ref{tab:top_comparison}). 
\item The performance gap between Random+FM and FM in Criteo dataset indicates that random selection on feature interactions may outperform the model keeping all the feature interactions under some circumstances, which supports our statement: removing some useless feature interactions could improve the performance.
\item The comparison between AutoFM and AutoFM-BN validates the effectiveness of BN in the \emph{re-train stage}, where the reason is stated in "AutoFIS" section.
\item The performance gap between AutoFM-BN and AutoFM-BN-$\alpha$ shows that $\boldsymbol{\alpha}$ improve the performance, as it differentiates the contribution of different feature interactions in the \emph{re-train stage}.
\end{enumerate}

\begin{table}[h]
    \centering
    % \tiny
    \vspace{-1em}
    \caption{Comparison of one-level and bi-level optimization}
    \vspace{-0.5em}
    \resizebox{0.38\textwidth}{!}{
   \begin{tabular}{c|cc|cc|}
\hline
\multicolumn{1}{|c|}{\multirow{2}{*}{Model}}& \multicolumn{2}{c|}{Avazu} & \multicolumn{2}{c|}{Criteo}\\
\cline{2-5}
\multicolumn{1}{|c|}{} & AUC & log loss  & AUC & log loss  \\
\hline
\multicolumn{1}{|c|}{AutoFM}  & \textbf{0.7831}  &\textbf{0.3778} &     \textbf{0.7974} & \textbf{0.5446}   \\ 
\multicolumn{1}{|c|}{Bi-AutoFM}  & 0.7816  &0.3787  &  0.7957  & 0.5464   \\ 
\hline
\multicolumn{1}{|c|}{AutoDeepFM}  &  \textbf{0.7852} & \textbf{0.3765} &   \textbf{0.8009} & \textbf{0.5404}  \\ 

\multicolumn{1}{|c|}{Bi-AutoDeepFM}  &  0.7843  & 0.3771  &  0.8002 & 0.5412     \\ 
\hline
\end{tabular}
}
\vspace{-1em}
\label{tab:opt-level}
\end{table}
\subsubsection{One-level V.S. bi-level optimization}

In this section, we compare the one-level and bi-level optimization on AutoFM and the results are presented in Table~\ref{tab:opt-level}. The performance gap between AutoFM and Bi-AutoFM (and between AutoDeepFM and Bi-AutoDeepFM) demonstrates the superiority of one-level optimization over bi-level, with the reason stated in "One-level Optimization" section. 

\section{CONCLUSION}
In this work, we proposed AutoFIS to automatically select important $2^{nd}$- and $3^{rd}$-order feature interactions. The proposed methods are generally applicable to all the \textit{factorization models} and the selected important interactions can be transferred to other deep learning models for CTR prediction. The proposed AutoFIS is easy to implement with marginal search costs, and the performance improvement is significant in two benchmark datasets and one private dataset. The proposed methods have been deployed onto the training platform of Huawei App Store recommendation service, with significant economic profit demonstrated.


\bibliographystyle{ACM-Reference-Format}
\bibliography{kdd}

\newpage
\appendix
\section{Parameter Settings}
For Avazu and Criteo datasets, the parameters of baseline models are set following~\cite{pin}. For AutoFM and AutoDeepFM we use the same hyper-parameters as the base models (i.e., FM and DeepFM accordingly) except for extra ones in AutoFM and AutoDeepFM.

\begin{center}
\begin{threeparttable}[t]
%\vspace{-1em}
\caption{Parameter Settings}
\label{tab:parameter}
\centering
\begin{tabular}{|c|l|l|}
\hline
Model  &  Avazu&Criteo  \\
\hline
General &  \begin{tabular}[c]{@{}l@{}}bs=2000\\ opt=Adam\\ lr=1e-3\end{tabular} &
\begin{tabular}[c]{@{}l@{}}bs=2000\\ opt=Adam\\ lr=1e-3\end{tabular}  \\
\hline
GBDT+LR&  \begin{tabular}[c]{@{}l@{}}\#tree=50\\\#child=2048\end{tabular} 
&  \begin{tabular}[c]{@{}l@{}}\#tree=80\\\#child=1024\end{tabular} 
\\\hline
GBDT+FFM &  \begin{tabular}[c]{@{}l@{}}\#tree=50\\\#child=1024\end{tabular} 
&  \begin{tabular}[c]{@{}l@{}}\#tree=20\\\#child=512 \end{tabular} 
\\\hline
FM &  \begin{tabular}[c]{@{}l@{}}k=40\end{tabular} &
 \begin{tabular}[c]{@{}l@{}}k=20\end{tabular}  \\\hline
FwFM &  \begin{tabular}[c]{@{}l@{}}k=40 \\wt\_init =0.7 \\ wt\_l1 = 1e-8 \\wt\_l2=1e-7 \end{tabular} &
 \begin{tabular}[c]{@{}l@{}}k=20\\wt\_init =0.7 \\ wt\_l1 = 0 \\wt\_l2=1e-7\end{tabular}  \\\hline
FFM & k=4 & k=4  \\ \hline
AFM &  \begin{tabular}[c]{@{}l@{}}k=40\\t=1\\h=256\\l2\_a =0 \end{tabular} &
 \begin{tabular}[c]{@{}l@{}}k=20\\t=0.01\\h=32\\l2\_a=0.1\end{tabular}  \\\hline

DeepFM & \begin{tabular}[c]{@{}l@{}}k=40\\net=[700$\times$ 5, 1]\\l2=0\\drop=1\\ BN=True\end{tabular} 
& \begin{tabular}[c]{@{}l@{}}k=20\\net=[700$\times$ 5, 1]\\l2=0\\drop=1\\ BN=True\end{tabular}
\\\hline
AutoDeepFM & \begin{tabular}[c]{@{}l@{}}c=0.0005\\mu=0.8\end{tabular} 
& \begin{tabular}[c]{@{}l@{}}c=0.0005\\mu=0.8\end{tabular} 
\\\hline
AutoFM &  \begin{tabular}[c]{@{}l@{}}c=0.005\\mu=0.6\end{tabular} 
&  \begin{tabular}[c]{@{}l@{}}c=0.0005\\mu=0.8 \end{tabular} 
\\\hline
\end{tabular}

\begin{tablenotes}
    \item[*] Note: bs=batch size, opt=optimizer, lr=learning rate,  k=embedding size, wt\_init = initial value for $ \alpha $, wt\_l1 = $l_1$ regularization on $\alpha$, wt\_l2 = $l_2$ regularization on $\alpha$, t=Softmax Temperature, l2\_a= L2 Regularization on Attention Network, net=MLP structure, LN=layer normalisation, BN=batch normaliation, c and mu are parameters in GRDA Optimizer.
    
    % drop\_a = dropout on attention network, h=attention network hidden size, n=embedding size, net=MLP structure, LN=layer normalisation, BN=batch normaliation, wt\_init = initial value for $ \alpha $, wt\_l1 = $l_1$ regularization on $\alpha$, wt\_l2 = $l_2$ regularization on $\alpha$, top= the ratio of selected $2^{nd}$ order feature interactions.
\end{tablenotes}
    \end{threeparttable}
    \end{center}

\end{document}